\documentclass[runningheads]{llncs}
\usepackage{graphicx}

\usepackage{tikz}
\usepackage{comment}
\usepackage{amsmath,amssymb} 
\usepackage{color}

\usepackage[accsupp]{axessibility}  

\usepackage{booktabs}
\usepackage{threeparttable}
\usepackage{multirow}
\usepackage{amsmath}

\begin{document}
\pagestyle{headings}
\mainmatter
\def\ECCVSubNumber{6680}  

\title{Improving Fine-Grained Visual Recognition in Low Data Regimes via Self-Boosting Attention Mechanism} 

\titlerunning{Self-Boosting Attention Mechanism}
%
\author{Yangyang Shu\inst{1}\and 
Baosheng Yu\inst{2} \and
Haiming Xu\inst{1} \and
Lingqiao Liu\inst{1}\thanks{Corresponding author: lingqiao.liu@adelaide.edu.au. This work is supported by the Centre for Augmented Reasoning.}}
\authorrunning{Y. et al.}
%
\institute{School of Computer Science, The University of Adelaide 
\email{\{yangyang.shu,hai-ming.xu,lingqiao.liu\}@adelaide.edu.au}\and
School of Computer Science, The University of Sydney 
\email{baosheng.yu@sydney.edu.au}\\}

\maketitle

\begin{abstract}

The challenge of fine-grained visual recognition often lies in discovering the key discriminative regions. While such regions can be automatically identified from a large-scale labeled dataset, a similar method might become less effective when only a few annotations are available. In low data regimes, a network often struggles to choose the correct regions for recognition and tends to overfit spurious correlated patterns from the training data. To tackle this issue, this paper proposes the self-boosting attention mechanism, a novel method for regularizing the network to focus on the key regions shared across samples and classes. Specifically, the proposed method first generates an attention map for each training image, highlighting the discriminative part for identifying the ground-truth object category. Then the generated attention maps are used as pseudo-annotations. The network is enforced to fit them as an auxiliary task. We call this approach the self-boosting attention mechanism (SAM). We also develop a variant by using SAM to create multiple attention maps to pool convolutional maps in a style of bilinear pooling, dubbed SAM-Bilinear. Through extensive experimental studies, we show that both methods can significantly improve fine-grained visual recognition performance on low data regimes and can be incorporated into existing network architectures. The source code is publicly available at: \textbf{\textit{\url{https://github.com/GANPerf/SAM}}}.


\end{abstract}

\section{Introduction}
Fine-Grained Visual Recognition (FGVR) aims to distinguish subcategories of objects under basic-level category, such as bird species~\cite{berg2014birdsnap,welinder2010caltech}, vehicle models~\cite{krause20133d,yang2015large}, aircraft models~\cite{maji2013fine}. The key challenge of FGVR is to discover the key object parts that can be used to identify object categories. In the existing works, such a discovery is either explicitly achieved through part-mining  \cite{he2017weakly,wei2018mask} or implicitly learned in end-to-end training \cite{min2020multi,zhuang2020learning}. The latter strategy is the current state-of-the-art, which usually relies on special designs of the network, e.g., bilinear networks \cite{lin2015bilinear,gao2016compact,kong2017low,zheng2019learning}, to impose certain inductive bias. 

Existing FGVR research is often based on a dataset with sufficient annotations, generally with more than 5,000 
images and hundreds of categories. However, many practical FGVR problems do not have such a large dataset since annotating fine-grained data is a time-consuming,
costly, and error-prone task. For example, labeling different bird species requires an expert in zoology. It remains unclear if the existing end-to-end learning methods can generalize well in the low data regime. 

Unfortunately, from our empirical study (shown in section \ref{CVS}), we found that the existing solutions for FGVR may become less effective. It seems that when the number of training samples becomes smaller, the network tends to overfit the spurious patterns that happen to correlate with object categories. For example, when distinguishing different types of birds, the network may pay more attention to the surrounding environment rather than the bird body.

To overcome this issue, in this paper, we propose a novel solution called the self-boosting attention mechanism (SAM) to regularize the network to make the decision based on regions that are shared across instances and categories. Specifically, we first use existing visual explanation approaches such as CAM \cite{zhou2016learning} and GradCAM \cite{selvaraju2017grad} to obtain attention maps to highlight the key regions supporting the prediction of the ground-truth class. Then we use the generated attention maps as prediction targets and fit them with a class-agnostic projection from the convolutional feature map. In this way, we could encourage the network to use the features from the commonly attended regions to make a prediction. To further strengthen the regularization, we further developed a variant by using the above auxiliary task to regularize a set of projections, with each projection working as a part detector. Those projections allow us to leverage a bilinear pooling operation to obtain a new representation of the image.
Through extensive experiments, we show that the proposed two strategies achieve superior performance than the competitive approaches for FGVR in a low data regime. Also, we demonstrate that the proposed method can be easily incorporated into the existing approach and achieve further performance boost.

\section{Related Work}

\subsection{Fine-grained Visual Recognition}
Locating distinctive regions plays an important and fundamental role in fine-grained visual recognition. In early researches, manually defined object and part annotations are extensively studied for fine-grained visual recognition. For example, Zhang~\emph{et al.}\cite{zhang2014part} use the trained R-CNN model to learn whole-object and part detectors with the help of part-level bounding boxes. Branson~\emph{et al.}\cite{branson2014bird} propose a method which is based on part detection. They use part and object bounding boxes to estimate a similarity-based warping function for improving the performance in fine-grained recognition tasks. However, manually defining object and part annotations requires additional human cost, largely limited in practical application. In the visual attention models community, Sermanet~\emph{et al.}\cite{sermanet2014attention} first propose to use attention models in FGVR. They use an attention-based RNN structure to direct high-resolution attention to the discriminative regions. However, the computational cost in their method is higher because they forward GoogLeNet three times. Xiao~\emph{et al.}\cite{xiao2015application} propose to use three types of attention in a deep neural network for the fine-grained classification task. The three types of attention are combined to train domain-specific deep nets: bottom-up attention, object-level top-down attention, and part-level top-down attention. Hu~\emph{et al}\cite{hu2019see} use weakly supervised learning to generate attention maps only by image-level annotation. The generated attention maps in their proposed WS-DAN network ensure the model looks at the object better and closer. The main drawback is that attention models will be vulnerable and prone to over-fitting when the image-level annotation is quite a few. 

Bilinear-based methods are very popular in fine-grained visual recognition. Lin~\emph{et al.}\cite{lin2015bilinear} first propose bilinear CNN models by two feature extractors to model local pairwise feature interactions for fine-grained visual recognition. Because original bilinear CNN models are high-dimensional and computationally expensive to train due to calculating pairwise interaction between channels, various studies of dimension reduction techniques have been proposed. Gao~\emph{et al.}\cite{gao2016compact} propose two compact bilinear pooling methods using two low-dimensional approximations of the polynomial kernel, Random Maclaurin\cite{kar2012random} and Tensor Sketch~\cite{pham2013fast} to generate the compact bilinear representations and reduce feature dimensions. Kong~\emph{et al.}\cite{kong2017low} present a compact low-rank classification model and use the low-rank approximation to the covariance matrix to address the computational demands of high feature dimensionality. Zheng~\emph{et al}\cite{zheng2019learning} propose a deep bilinear transformation (DBT) block to uniformly divide input channels into several semantic groups. The computational cost can be relieved via calculating pairwise interactions within each group. For our method, the feature dimensions can be reduced via controlling the number of the predicted attention maps when the element-wise multiplication.

\subsection{Low-Supervised FGVR}
To reduce the dependence on training data, some studies distinguish different categories with very little supervision, e.g., few-shot fine-grained visual recognition and semi-supervised learning for fine-grained visual recognition. Zhu~\emph{et al.}\cite{zhu2020multi} propose a multi-attention meta-learning (MattML) method to capture discriminative parts of images for few-shot FGVR. The proposed MattML consists of the base learner and task learner, where the base learner is used for general feature learning, and the task learner uses a task embedding network to learn task representations. Wei~\emph{et al.}\cite{wei2019piecewise} proposed an end-to-end trainable deep network to solve few-shot FGVR. They use a bilinear feature learning module to capture the discriminative information of an exemplar image and use a classifier mapping module to map the intermediate feature into the decision boundary of the novel category. Lai~\emph{et al.}\cite{lai2019improving} propose an efficient method of semi-supervised learning, voted pseudo label (VPL), to improve the performance of classification in FGVR task when only a few samples are available. VPL is applied in unlabeled data to pick up their classes with non-confused labels, verified by the consensus prediction of different classification models. Mugnai~\emph{et al}\cite{mugnai2022fine} exploit semi-supervised learning to improve the performance of FGVR. They adopt an adversarial optimization strategy to combine the conditional entropy of unlabeled data with a second-order feature encoder to reduce the prohibitive annotation cost of FGVR. Our method works in a different setting to the above methods. Specifically, unlike semi-supervised FGVR, we do not assume the availability of unlabeled data; unlike few-shot FGVR, we do not require a relatively large amount of labeled samples from different categories as ``base class samples''. 

\section{Background}

In this section, we briefly review Class Activation Maps (CAM)~\cite{zhou2016learning} and Gradient-weighted Class Activation Mapping (Grad-CAM)~\cite{selvaraju2017grad}, which underpins the proposed Self-boosting Attention Mechanism.

\subsection{Class Activation Maps}
Class Activation Maps (CAM) are proposed to identify the importance of the image regions by projecting back the
weights of the output layer onto the convolutional feature maps. CAM is applicable for the neural network architecture that uses Global Average Pooling (GAP) layer and classifier layers as the last two layers.

Let $\phi(I) \in \mathbb{R}^{H \times W \times D}$ represents the activation feature map of the last convolutional layer, where $I$ is the input image and $H$, $W$ and $D$ are the height, width and the number of channels of the feature map, respectively. Thus the logits for class $y$, i.e., the decision value before the softmax, can be calculated as:
\begin{equation}
\begin{split}
\label{cam1}
l(y) &= \mathbf{w_y}^\top GAP(\phi(I))= \mathbf{w_y}^\top \frac{1}{HW} \sum_{i=1}^H \sum_{j=1}^W [\phi(I)]_{i,j}=\frac{1}{H W}\sum_{i=1}^H \sum_{j=1}^W \mathbf{w_y}^\top [\phi(I)]_{i,j},
\end{split}
\end{equation}
where $\mathbf{w_y}$ is the classifier for the $y$-th class. $GAP$ represents Global Average Pooling and $[\phi(I)]_{i,j} \in \mathbb{R}^D$ denotes the feature vector located at the $(i,j)$-th grid. The class activation map (CAM) for class $y$ is defined as:
\begin{equation}
\begin{split}
\label{cam2}
[\mathrm{CAM}(y)]_{i,j} = \mathbf{w_y}^\top [\phi(I)]_{i,j},
\end{split}
\end{equation}where $\mathrm{CAM}(y)$ denotes CAM for the $y$-th class. $[\mathrm{CAM}(y)]_{i,j}$ indicates the importance value of the $(i,j)$th spatial grid.

\subsection{Gradient-weighted Class Activation Mapping}
Gradient-weighted class activation mapping (GradCAM) extends CAM by using the gradient information to calculate the importance of the activation. Unlike CAM, GradCAM could be applied to any convolutional layer or the input image, and is applicable to any neural network architecture. Formally, the GradCAM for the $y$-th class is calculated via:
\begin{equation}
\begin{split}
\label{gradcam}
[\mathrm{Grad\text{-}CAM}(y)]_{i,j} =  ReLU\left([\frac{\partial l(y)}{\partial [\phi(I)]_{i,j}}]^\top[\phi(I)]_{i,j}\right),
\end{split}
\end{equation}
where $\mathrm{Grad\text{-}CAM}(y)$ denotes GradCAM for the $y$-th
class. $[\mathrm{Grad\text{-}CAM}(y)]_{i,j}$ refers to the importance value of the $(i,j)$th spatial grid. $l(y)$ is the logits for class $y$. $\frac{\partial l(y)}{\partial \phi(I)}$ is the gradient of the logits for class $y$ w.r.t. the feature map $\phi(I)$.

\section{Our Methods}
\begin{figure}[t]
	\centering
	\includegraphics[width=4in]{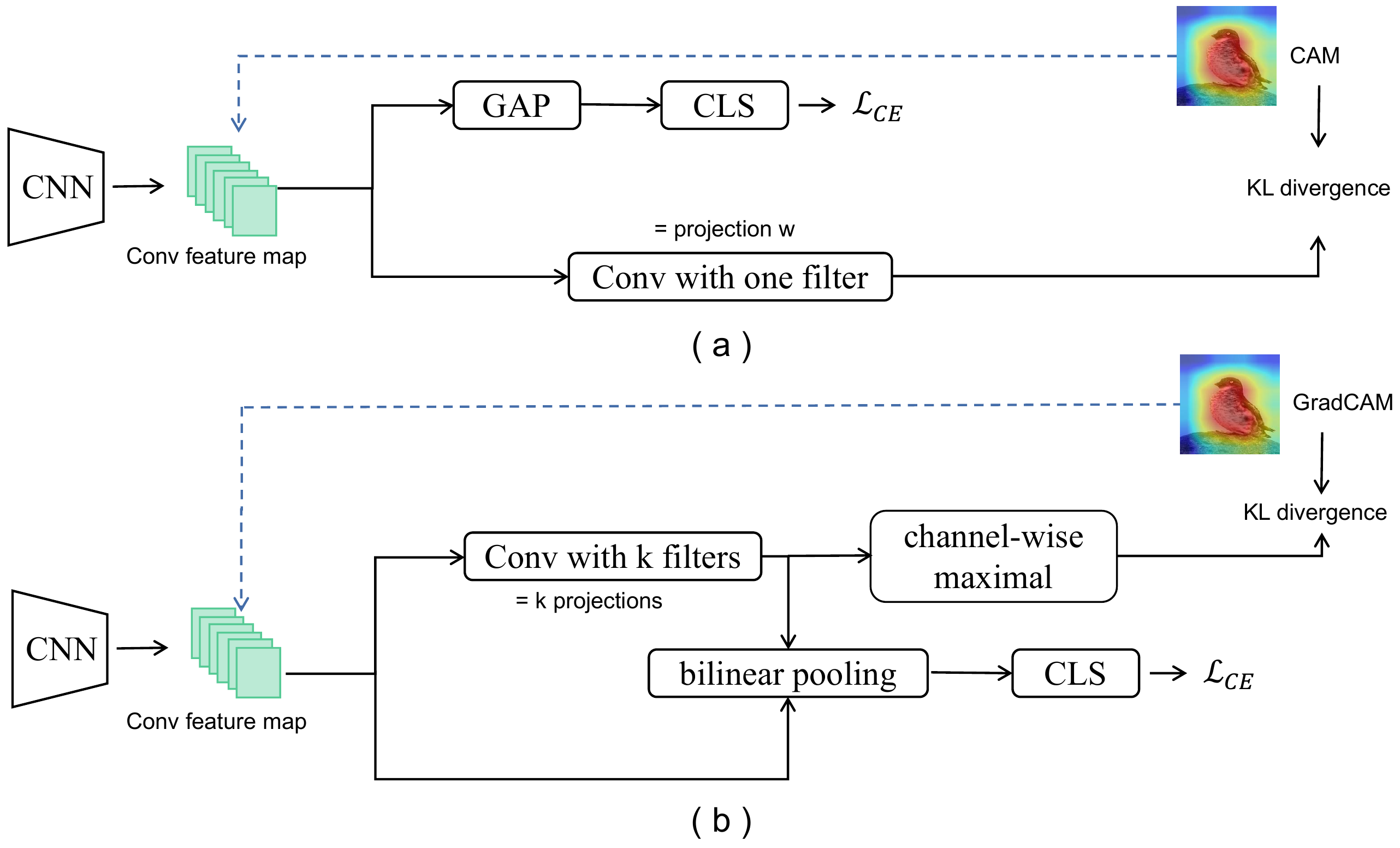}
	\caption{The overview of our SAM network architecture. In the top half-section (a), the last convolutional layer feature maps in CNN network are used to obtain the cross-entropy loss function $\mathcal{L}_{CE}$ via global average pooling (GAP) and classifier (CLS). These feature maps via a linear projection are also enforced to fit the attention maps generated by CAM. The bottom half section (b) represents the method developed in bilinear pooling. The multiple projections are applied in convolutional feature maps to obtain multiple part detectors. Then a bilinear pooling operation is used to obtain a new feature representation. The feature maps after multi-projection and channel-wise maximal operation are also enforced to fit the attention maps generated by GradCAM.}
	\label{framework}
\end{figure}
\subsection{Self-boosting Attention Mechanism}
\label{method_sam}
As introduced in the Introduction, the challenge of a fine-grained visual recognition system is to identify the key regions that can be discriminative for distinguishing the subtle difference across categories. With abundant training data and properly designed architecture, the key regions can usually be automatically learned via end-to-end training. However, as shown in our experiment (please see section \ref{AAnalysis} and \ref{CVS}), such an end-to-end training strategy becomes less effective in identifying the key regions when the number of training samples becomes smaller. In such a case, the spurious correlation and true discriminative patterns become hard to distinguish, and a network often mistakenly utilizes the former, leading to poor generalization.

To overcome this issue, this paper proposes a self-boosting strategy to regularize the network to encourage the use of regions shared across many instances and classes. 
Following the above notation, we hereafter use $\phi(I) \in \mathbb{R}^{H\times W\times D}$ to denote the last convolutional layer feature maps. 
The logits is obtained by applying a classifier $h_p(\cdot)$ to $\phi(I)$, that is, $p(y|I) = h_p(GAP(\phi(I)))$. The proposed regularization strategy constructs an auxiliary task for $\phi(I)$. Specifically, we first calculate CAM or GradCAM as attention maps\footnote{Once calculated, the attention map is detached from the back-propagation.} from $\phi(I)$ w.r.t the ground-truth class for each instance, denoting $g(I_n,y_n)$. Then we enforce $\phi(I)$ to fit $g(I_n,y_n)$ via a linear projection $\mathbf{w} \in \mathbb{R}^D$ without providing the ground-truth class information to the network, which could be implemented as applying a convolutional layer with a single filter. Specifically, we first normalize $g(I_n,y_n)$ and $\mathbf{w}^T\phi(I) \in \mathbb{R}^{H \times W}$ via the softmax function:
\begin{align}\label{eq: normalization}
    \mathbf{\bar{G}} = \frac{\exp{(g(I_n,y_n)/\tau)}}{\sum_{i=1}^H \sum_{j=1}^W \exp{{([g(I_n,y_n)]_{i,j}/\tau)}}} \in \mathbb{R}^{H \times W}, \nonumber \\
    \mathbf{\bar{A}} = \frac{\exp{(\mathbf{w}^T\phi(I)/\tau)}}{\sum_{i=1}^H \sum_{j=1}^W \exp{{([\mathbf{w}^T\phi(I)]_{i,j}/\tau)}}} \in \mathbb{R}^{H \times W},
\end{align}where $[\cdot]_{i,j}$ denotes the $i,j$-th element of the feature map. $\tau$ is an empirical temperature parameter and we set it to 0.4 for all our experiments. For the simplistic of notations, we also slightly abuse the notation $\mathbf{w}^T\phi(I)$ to denote the feature map obtained by projecting the vector at each location of $\phi(I)$ through $\mathbf{w}$. This normalization will highlight the most important regions and we can also view the normalized feature maps are probability distributions. Then we can use Kullback-Leibler divergence, denoted as $\mathrm{KL}(\cdot,\cdot)$ to measure the compatibility between $g'$ and $p'$. Thus the final loss function could be written as 
\begin{align}
    \mathcal{L} = \mathcal{L}_{CE} + \lambda \mathcal{L}_{SAM} =  \mathcal{L}_{CE} + \lambda \mathrm{KL}(vec(\mathbf{\bar{A}}),vec(\mathbf{\bar{G}})).
\end{align}

At the first glance, the introduction of $\mathcal{L}_{SAM}$ seems to be slightly counter-intuitive. The attention map is generated from the current model, why allowing model to fit it will lead to any benefit? To understand its effect, one should notice that the attention map is calculated based on the ground-truth class. For example, if we use CAM to calculate the attention map, the CAM is calculated via $\mathbf{w}_{y_n}^\top [\phi(I_n)]_{i,j}$, that is, the classifier corresponding to the ground-truth class $y_n$ is chosen to produce the CAM. In contrast, the projection vector $\mathbf{w}$ is class-agnostic. Thus, $\mathbf{w}^\top \phi(I)$ tends to fit the common part that are shared across all classes and instances. Also in this process, $\phi(I)$ will be learned to produce a good feature presentation for those common key parts. This in effect creates an inductive bias for encouraging the network to use the patterns from the common key parts to make prediction. We call this mechanism as self-boosting attention mechanism since the model will be boosted by fitting its own attention map. The illustration of this scheme can be seen in Figure \ref{framework} (a).

\subsection{A Bilinear Pooling Extension of SAM} 
\label{extension_sam_bilinear}
The rationale of the aforementioned SAM is that if $\phi(I)$ is learned to support detecting common parts via the auxiliary task, the network will also prefer to use the feature from the common parts to make a prediction. In such a design, we do not have hard constraints to enforce the network to only use features extracted from those common parts. In this section, we present an extension of SAM by explicitly introducing such a constraint.

The most straightforward approach is to use $A = \mathbf{w}^\top \phi(I)$ as an attention map to weight $\phi(I)$. In other words, instead of directly applying global average pooling to $\phi(I)$ to obtain image representation, we use the following attentive pooling scheme:
\begin{align}\label{eq:attention pooling}
    \mathbf{f} = \sum_{i,j} [A]_{i,j} [\phi(I)]_{i,j} \in \mathbb{R}^D.  
 \end{align} With such a pooling scheme, the feature from un-attended regions, i.e. $[A]_{i,j} = 0$, will not be preserved into the pooled representation.

We notice that using attentive pooling is akin to the operation in bilinear pooling while the later is equivalent to using multiple attentions. Inspired by this analogy, we further create multiple attention maps $\{A_k|A_k = \mathbf{w}_k^\top \phi(I)\}, k = 1\cdots K$ via multiple projections $\{\mathbf{w}_k\}$. Intuitively, we expect each attention map highlights one object part, and the union of them should fit the attention map calculated from GradCAM for the current image, as the latter showing all important regions contributing to the decision. In our method, we approximate the union of identified object key parts via taking the maximal value across all $K$ attention maps, that is,
 \begin{align}\label{eq:channelwise max-pooling}
    [A_u]_{i,j} = \max_k [A_k]_{i,j}
 \end{align}

Then for each attention map, we can create a pooled feature via Eq. \ref{eq:attention pooling}. We then concatenate the pooled features from all $K$ feature maps as the final image representation:
\begin{align}\label{eq:bilinear SAM}
     &\mathbf{f} = cat(\mathbf{f}_1,\mathbf{f}_2,\cdots,\mathbf{f}_K) \in \mathbb{R}^{DK} \nonumber \\
     &\mathbf{f}_k = \sum_{i,j} [A_k]_{i,j} [\phi(I)]_{i,j},
\end{align} where $cat()$ denotes concatenation of vectors. 
Note that Eq. \ref{eq:bilinear SAM} is identical to the bilinear pooling \cite{lin2017bilinear}. So we call this extension SAM-Bilinear. 

The above idea can be implemented by following a bilinear neural network structure. The illustration of this scheme is shown in Figure \ref{framework} (b). To summarize, the network introduces a bilinear pooling module with one input being the last convolutional layer feature map $\phi(I)$ and the other input being $K$ projections of $\mathbf{w}_k^\top \phi(I)$. This could be implemented by adding a convolutional layer with $K$ filters after $\phi(I)$. The attention map is calculated by using GradCAM with respect to $\phi(I)$. Then a channel-wise max-pooling (Eq. \ref{eq:channelwise max-pooling}) is applied after the added convolutional layer to obtain a predicted attention map. We then normalized both the predicted attention map and the generated attention map by following the scheme in Eq. \ref{eq: normalization}.

\section{Experiments}
In this section, we conduct experiments to evaluate the performance of the SAM model for FGVR. The experimental conditions, including datasets, implementation Details etc. are firstly given in Section~\ref{ECon}. In Section~\ref{AAnalysis}, ablation analysis are performed to investigate the effectiveness of each component in our model. We give the comparison with state-of-the-art methods in Section~\ref{CVS}. In Section~\ref{attentionnumber}, we conduct experiments to exploit the effect of the number of line projections. Lastly, we use the visualization to explain our model in Section~\ref{Vis}.
\subsection{Experimental Conditions}
\label{ECon}
\subsubsection{Datasets}
\begin{table}[tbp]
\caption{Category and data splits on the \texttt{CUB-200-2011}, \texttt{Stanford Cars} and \texttt{FGVC-Aircraft} datasets}
	\label{dataset}
	\centering
\begin{tabular}{|c|c|c|c|}
\hline
Datasets      & Category & No. of training & No.of testing \\ \hline
\texttt{CUB-200-2011}  & 200      & 5994            & 5794          \\
\texttt{Stanford Cars} & 196      & 8144            & 8041          \\
\texttt{FGVC-Aircraft} & 100      & 6667            & 3333          \\ \hline
\end{tabular}
\end{table}
In our experiments, we use three publicly available fine-grained visual datasets: \texttt{Caltech-UCSD Birds (CUB-200-2011)}~\cite{wah2011caltech}, \texttt{Stanford Cars}~\cite{krause20133d} and \texttt{FGVC-Aircraft}~\cite{maji2013fine}. The details of category and data splits in these three datasets are shown in Table~\ref{dataset}. We reduce the number of labeled annotations, i.e., $10\%$ to $50\%$ for each category and the number of categories in our experiments to simulate the scenarios of low data regimes.

\subsubsection{Implementation Details}
We implement our method using the PyTorch framework. In our experiments, the input images are resized to 256$\times$256. Then a 224$\times$224 patch is cropped randomly from the rescaled images on the three datasets for the purpose of data augmentation. ResNet-50~\cite{he2016deep} is used as the architecture, and layer four is chosen as feature maps. The pre-trained weights of ResNet-50 on Imagenet are used for initialization. SGD optimizer with a mini-batch size of 24, weight decay of $1\times10^{-4}$, and a momentum of 0.9 are used to optimize the proposed network in our experiments. The learning rate of the classifier is 0.001. The parameter $\lambda$ is 0.01.
\subsubsection{Baselines}
We now compare our method to bilinear pooling-based methods. We choose the following four popular methods for comparison. For a fair comparison, we re-implement the method by changing VGG~\cite{simonyan2014very} with the ResNet framework.
\begin{itemize}
    \item \textbf{Full Bilinear Pooling (FBP)~\cite{lin2017bilinear}} uses an image as the input of two CNNs, and their outputs at each location are combined to obtain the bilinear feature representation. In~\cite{lin2017bilinear}, the $relu5_{\_}3$ layer and $relu5$ layer truncated in a VGG-D~\cite{simonyan2014very} and VGG-M~\cite{chatfield2014return} networks respectively are used for obtaining bilinear. In this paper, we re-implement the method by following the identical structures as SAM-Bilinear for fair comparison.  Specifically, we truncate at layer four of ResNet framework and apply $K$ projections into the truncated layer four. The last convolutional feature maps and the outputs of projections are used to obtain the bilinear feature representation.
    \item \textbf{Compact Bilinear Pooling (CBP-TS)~\cite{gao2016compact}} with Tensor Sketch projection is used in the same extract experimental setup as FBP. The projection dimension in~\cite{gao2016compact} is found as $d=8000$ to reach close-to maximum accuracy. We set d=500 to reach the maximum accuracy in our experiments. 
    \item \textbf{Hierarchical Bilinear Pooling (HBP)~\cite{yu2018hierarchical}} integrates multiple cross-layer bilinear features to improve their representation capability. $relu5_{\_}1$, $relu5_{\_}2$ and $relu5_{\_}3$ in VGG-16 in \cite{yu2018hierarchical} are used because deeper layers contain more part semantic information. This paper applies HBP to ResNet network structures in the deeper layers (layer four). The dimension of joint embedding $D$ in~\cite{yu2018hierarchical} is 8192*3. Given the computational complexity and classification performance, we set the same value as HBP in ResNet network structures.
    \item \textbf{Deep Bilinear Transformation (DBTNet-50)~\cite{zheng2019learning}} divides input channels into several semantic groups according to their semantic information and calculates pairwise interaction within semantic groups to obtain bilinear features efficiently. This also results in large saving in computation cost.
\end{itemize}
\subsubsection{Experimental Design}
To meet the situation of a few annotations, we set the image-level annotations with four ratios, i.e., $10\%$, $15\%$, $30\%$, and $50\%$. Although only a few annotations are available, the proposed method employs a self-boosting attention mechanism to regularize the network and improve the classification performance of fine-grained tasks. 


\subsection{Main Results}
\label{AAnalysis}
\begin{table}[htbp]
\caption{Evaluation of our method with four label proportions and four label categories on the
\texttt{CUB200-2011} (Bird), \texttt{Stanford Cars} (Car) and \texttt{FGVC aircraft} (Aircraft) databases. SAM ResNet-50 applies the proposed SAM to the Resnet-50. Similarly, SAM-Bilinear combines the proposed SAM with FBP. (Bold numbers indicate the best performance. $\uparrow$ is the amount of increase compared to the respective baseline of SAM and SAM-Bilinear, i.e., ResNet-50 for SAM and FBP for SAM-Bilinear.)}
\label{ablation_3datasets}
	\centering
	\resizebox{\textwidth}{80mm}{
\begin{tabular}{|c|c|l|llll|}
\hline
\multirow{2}{*}{Dataset}   & \multirow{2}{*}{Category} & \multicolumn{1}{c|}{\multirow{2}{*}{Method}} & \multicolumn{4}{c|}{Label Proportion}                                                                                                                                                                                                     \\ \cline{4-7} 
                           &                           & \multicolumn{1}{c|}{}                        & \multicolumn{1}{c|}{10\%}                                     & \multicolumn{1}{c|}{15\%}                                      & \multicolumn{1}{c|}{30\%}                                     & \multicolumn{1}{c|}{50\%}                \\ \hline\hline
\multirow{16}{*}{Bird}     & \multirow{4}{*}{30}       & ResNet-50                                    & \multicolumn{1}{l|}{55.56\%}                                  & \multicolumn{1}{l|}{61.55\%}                                   & \multicolumn{1}{l|}{69.16\%}                                  & 77.65\%                                  \\
                           &                           & SAM ResNet-50                                         & \multicolumn{1}{l|}{58.76\%\tiny $\uparrow$3.20\%}                           & \multicolumn{1}{l|}{\textbf{65.79\%}\tiny $\uparrow$4.24\%}  & \multicolumn{1}{l|}{70.16\%\tiny $\uparrow$1.00\%}                           & 77.75\%\tiny $\uparrow$0.10\%                           \\
                           &                           & FBP                                    & \multicolumn{1}{l|}{56.55\%}                                  & \multicolumn{1}{l|}{61.93\%}                                   & \multicolumn{1}{l|}{69.79\%}                                  & 77.86\%                                  \\
                           &                           & SAM bilinear                               & \multicolumn{1}{l|}{\textbf{60.18\%}\tiny $\uparrow$3.63\%} & \multicolumn{1}{l|}{65.65\%\tiny $\uparrow$3.72\%}                            & \multicolumn{1}{l|}{\textbf{70.79\%}\tiny $\uparrow$1.00\%} & \textbf{78.28\%}\tiny $\uparrow$0.42\% \\ \cline{2-7} 
                           & \multirow{4}{*}{50}       &  ResNet-50                                   & \multicolumn{1}{l|}{45.72\%}                                  & \multicolumn{1}{l|}{58.53\%}                                   & \multicolumn{1}{l|}{68.47\%}                                  & 73.94\%                                  \\
                           &                           & SAM ResNet-50                                          & \multicolumn{1}{l|}{\textbf{51.14\%}\tiny $\uparrow$5.42\%} & \multicolumn{1}{l|}{61.80\%\tiny $\uparrow$3.27\%}                            & \multicolumn{1}{l|}{71.43\%\tiny$\uparrow$2.76\%}                           & 75.19\%\tiny$\uparrow$1.25\%                           \\
                           &                           & FBP                                    & \multicolumn{1}{l|}{42.82\%}                                  & \multicolumn{1}{l|}{58.24\%}                                   & \multicolumn{1}{l|}{68.67\%}                                  & 74.37\%                                  \\
                           &                           & SAM bilinear                               & \multicolumn{1}{l|}{50.46\%\tiny $\uparrow$7.64\%}                           & \multicolumn{1}{l|}{\textbf{61.92\%}\tiny $\uparrow$3.68\%}  & \multicolumn{1}{l|}{\textbf{72.07\%}\tiny $\uparrow$3.40\%} & \textbf{77.18\%}\tiny $\uparrow$2.81\% \\ \cline{2-7} 
                           & \multirow{4}{*}{100}      &  ResNet-50                                     & \multicolumn{1}{l|}{42.18\%}                                  & \multicolumn{1}{l|}{56.28\%}                                   & \multicolumn{1}{l|}{67.85\%}                                  & 75.39\%                                  \\
                           &                           & SAM ResNet-50                                         & \multicolumn{1}{l|}{46.27\%\tiny $\uparrow$4.09\%}                           & \multicolumn{1}{l|}{\textbf{60.16\%}\tiny $\uparrow$3.88\%}  & \multicolumn{1}{l|}{70.82\%\tiny $\uparrow$ 2.97\%}                           & 78.25\%\tiny $\uparrow$2.86\%                           \\
                           &                           & FBP                                   & \multicolumn{1}{l|}{42.52\%}                                  & \multicolumn{1}{l|}{55.94\%}                                   & \multicolumn{1}{l|}{68.85\%}                                  & 75.71\%                                  \\
                           &                           & SAM bilinear                               & \multicolumn{1}{l|}{\textbf{47.07\%}\tiny $\uparrow$4.55\%} & \multicolumn{1}{l|}{59.60\%\tiny$\uparrow$3.66\%}                            & \multicolumn{1}{l|}{\textbf{70.98\%}\tiny $\uparrow$2.13\%} & \textbf{78.53\%}\tiny $\uparrow$2.82\% \\ \cline{2-7} 
                           & \multirow{4}{*}{200}      &  ResNet-50                                    & \multicolumn{1}{l|}{36.99\%}                                  & \multicolumn{1}{l|}{48.88\%}                                   & \multicolumn{1}{l|}{62.60\%}                                  & 73.23\%                                  \\
                           &                           & SAM ResNet-50                                           & \multicolumn{1}{l|}{40.24\%\tiny $\uparrow$3.25\%}                           & \multicolumn{1}{l|}{52.05\%\tiny $\uparrow$3.17\%}                            & \multicolumn{1}{l|}{64.07\%\tiny $\uparrow$1.47\%}                           & 73.92\%\tiny$\uparrow$0.69\%                           \\
                           &                           & FBP                                    & \multicolumn{1}{l|}{37.88\%}                                  & \multicolumn{1}{l|}{49.12\%}                                   & \multicolumn{1}{l|}{63.27\%}                                  & 73.70\%                                  \\
                           &                           & SAM bilinear                               & \multicolumn{1}{l|}{\textbf{41.83\%}\tiny$\uparrow$3.95\%} & \multicolumn{1}{l|}{\textbf{52.35\%}\tiny$\uparrow$3.23\%}  & \multicolumn{1}{l|}{\textbf{65.19\%}\tiny$\uparrow$1.92\%} & \textbf{74.54\%}\tiny$\uparrow$0.84\% \\ \hline
\multirow{16}{*}{Car}      & \multirow{4}{*}{30}       &  ResNet-50                                     & \multicolumn{1}{l|}{35.09\%}                                  & \multicolumn{1}{l|}{45.72\%}                                   & \multicolumn{1}{l|}{58.65\%}                                  & 68.53\%                                  \\
                           &                           & SAM ResNet-50                                           & \multicolumn{1}{l|}{39.95\%\tiny$\uparrow$4.86\%}                           & \multicolumn{1}{l|}{49.98\%\tiny$\uparrow$4.26\%}                            & \multicolumn{1}{l|}{61.90\%\tiny$\uparrow$3.25\%}                           & \textbf{75.86\%}\tiny$\uparrow$2.33\% \\
                           &                           & FBP                                   & \multicolumn{1}{l|}{36.24\%}                                  & \multicolumn{1}{l|}{46.14\%}                                   & \multicolumn{1}{l|}{62.98\%}                                  & 73.92\%                                  \\
                           &                           & SAM bilinear                               & \multicolumn{1}{l|}{\textbf{41.76\%}\tiny$\uparrow$5.52\%} & \multicolumn{1}{l|}{\textbf{50.49\%}\tiny$\uparrow$4.35\%}  & \multicolumn{1}{l|}{\textbf{66.89\%}\tiny$\uparrow$3.91\%} & 75.37\%\tiny$\uparrow$1.43\%                           \\ \cline{2-7} 
                           & \multirow{4}{*}{50}       &  ResNet-50                                     & \multicolumn{1}{l|}{34.38\%}                                  & \multicolumn{1}{l|}{45.32\%}                                   & \multicolumn{1}{l|}{62.64\%}                                  & 76.67\%                                  \\
                           &                           & SAM ResNet-50                                          & \multicolumn{1}{l|}{42.39\%\tiny$\uparrow$8.01\%}                           & \multicolumn{1}{l|}{\textbf{54.23\%}\tiny$\uparrow$8.91\%}  & \multicolumn{1}{l|}{69.00\%\tiny$\uparrow$6.36\%}                           & 79.14\%\tiny$\uparrow$2.47\%                           \\
                           &                           & FBP                                   & \multicolumn{1}{l|}{37.76\%}                                  & \multicolumn{1}{l|}{44.53\%}                                   & \multicolumn{1}{l|}{63.43\%}                                  & 77.27\%                                  \\
                           &                           & SAM bilinear                              & \multicolumn{1}{l|}{\textbf{43.23\%}\tiny$\uparrow$5.47\%} & \multicolumn{1}{l|}{54.18\%\tiny$\uparrow$9.65\%}                            & \multicolumn{1}{l|}{\textbf{69.15\%}\tiny$\uparrow$5.72\%} & \textbf{79.40\%}\tiny$\uparrow$2.13\% \\ \cline{2-7} 
                           & \multirow{4}{*}{100}      &  ResNet-50                                     & \multicolumn{1}{l|}{36.56\%}                                  & \multicolumn{1}{l|}{47.46\%}                                   & \multicolumn{1}{l|}{69.77\%}                                  & 79.86\%                                  \\
                           &                           & SAM ResNet-50                                           & \multicolumn{1}{l|}{47.42\%\tiny$\uparrow$10.86\%}                          & \multicolumn{1}{l|}{\textbf{59.18\%}\tiny$\uparrow$11.72\%} & \multicolumn{1}{l|}{75.75\%\tiny$\uparrow$5.98\%}                           & 84.96\%\tiny$\uparrow$5.10\%                           \\
                           &                           & FBP                                  & \multicolumn{1}{l|}{38.55\%}                                  & \multicolumn{1}{l|}{50.32\%}                                   & \multicolumn{1}{l|}{71.96\%}                                  & 81.51\%                                  \\
                           &                           & SAM bilinear                               & \multicolumn{1}{l|}{\textbf{47.69\%}\tiny$\uparrow$9.14\%} & \multicolumn{1}{l|}{58.74\%\tiny$\uparrow$8.42\%}                            & \multicolumn{1}{l|}{\textbf{76.86\%}\tiny$\uparrow$4.9\%}  & \textbf{85.23\%}\tiny$\uparrow$3.72\% \\ \cline{2-7} 
                           & \multirow{4}{*}{196}      &  ResNet-50                                     & \multicolumn{1}{l|}{37.45\%}                                  & \multicolumn{1}{l|}{53.01\%}                                   & \multicolumn{1}{l|}{75.26\%}                                  & 83.56\%                                  \\
                           &                           & SAM ResNet-50                                         & \multicolumn{1}{l|}{39.96\%\tiny$\uparrow$2.51\%}                           & \multicolumn{1}{l|}{55.02\%\tiny$\uparrow$2.01\%}                            & \multicolumn{1}{l|}{76.69\%\tiny$\uparrow$1.43\%}                           & 84.85\%\tiny$\uparrow$1.29\%                           \\
                           &                           & FBP                                   & \multicolumn{1}{l|}{40.13\%}                                  & \multicolumn{1}{l|}{55.07\%}                                   & \multicolumn{1}{l|}{76.42\%}                                  & 85.10\%                                  \\
                           &                           & SAM bilinear                               & \multicolumn{1}{l|}{\textbf{43.19\%}\tiny$\uparrow$3.06\%} & \multicolumn{1}{l|}{\textbf{57.42\%}\tiny$\uparrow$2.35\%}  & \multicolumn{1}{l|}{\textbf{77.63\%}\tiny$\uparrow$1.21\%} & \textbf{85.71\%}\tiny$\uparrow$0.61\% \\ \hline
\multirow{12}{*}{Aircraft} & \multirow{4}{*}{30}       &  ResNet-50                                    & \multicolumn{1}{l|}{26.70\%}                                  & \multicolumn{1}{l|}{33.50\%}                                   & \multicolumn{1}{l|}{47.00\%}                                  & 63.00\%                                  \\
                           &                           & SAM ResNet-50                                          & \multicolumn{1}{l|}{31.80\%\tiny$\uparrow$5.10\%}                           & \multicolumn{1}{l|}{37.70\%\tiny$\uparrow$4.20\%}                            & \multicolumn{1}{l|}{49.15\%\tiny$\uparrow$2.15\%}                           & 65.10\%\tiny$\uparrow$2.10\%                           \\
                           &                           & FBP                                    & \multicolumn{1}{l|}{26.90\%}                                  & \multicolumn{1}{l|}{33.60\%}                                   & \multicolumn{1}{l|}{46.70\%}                                  & 61.90\%                                  \\
                           &                           & SAM bilinear                               & \multicolumn{1}{l|}{\textbf{32.50\%}\tiny$\uparrow$5.60\%} & \multicolumn{1}{l|}{\textbf{39.20\%}\tiny$\uparrow$5.60\%}  & \multicolumn{1}{l|}{\textbf{51.80\%}\tiny$\uparrow$5.10\%} & \textbf{65.80\%}\tiny$\uparrow$3.90\% \\ \cline{2-7} 
                           & \multirow{4}{*}{50}       &  ResNet-50                                   & \multicolumn{1}{l|}{38.60\%}                                  & \multicolumn{1}{l|}{45.20\%}                                   & \multicolumn{1}{l|}{61.16\%}                                  & 70.29\%                                  \\
                           &                           & SAM ResNet-50                                          & \multicolumn{1}{l|}{43.58\%\tiny$\uparrow$4.98\%}                           & \multicolumn{1}{l|}{49.88\%\tiny$\uparrow$4.68\%}                            & \multicolumn{1}{l|}{63.79\%\tiny$\uparrow$2.63\%}                           & 72.25\%\tiny$\uparrow$1.96\%                           \\
                           &                           & FBP                                    & \multicolumn{1}{l|}{37.94\%}                                  & \multicolumn{1}{l|}{45.44\%}                                   & \multicolumn{1}{l|}{61.48\%}                                  & 71.79\%                                  \\
                           &                           & SAM bilinear                               & \multicolumn{1}{l|}{\textbf{43.70\%}\tiny$\uparrow$5.76\%} & \multicolumn{1}{l|}{\textbf{50.84\%}\tiny$\uparrow$5.40\%}  & \multicolumn{1}{l|}{\textbf{65.33\%}\tiny$\uparrow$3.85\%} & \textbf{72.95\%}\tiny$\uparrow$1.16\% \\ \cline{2-7} 
                           & \multirow{4}{*}{100}      &  ResNet-50                                     & \multicolumn{1}{l|}{43.52\%}                                  & \multicolumn{1}{l|}{53.17\%}                                   & \multicolumn{1}{l|}{71.32\%}                                  & 78.61\%                                  \\
                           &                           & SAM ResNet-50                                         & \multicolumn{1}{l|}{46.73\%\tiny$\uparrow$2.21\%}                           & \multicolumn{1}{l|}{56.02\%\tiny$\uparrow$2.85\%}                            & \multicolumn{1}{l|}{72.59\%\tiny$\uparrow$1.27\%}                           & 79.21\%\tiny$\uparrow$0.60\%                           \\
                           &                           & FBP                                    & \multicolumn{1}{l|}{45.16\%}                                  & \multicolumn{1}{l|}{55.06\%}                                   & \multicolumn{1}{l|}{72.12\%}                                  & 79.93\%                                  \\
                           &                           & SAM bilinear                               & \multicolumn{1}{l|}{\textbf{47.97\%}\tiny$\uparrow$2.81\%} & \multicolumn{1}{l|}{\textbf{57.47\%}\tiny$\uparrow$2.41\%}  & \multicolumn{1}{l|}{\textbf{73.43\%}\tiny$\uparrow$1.31\%} & \textbf{80.86\%}\tiny$\uparrow$0.93\% \\ \hline
\end{tabular}}
\end{table}
To thoroughly investigate the proposed method, we conduct experiments to provide a detailed ablation analysis with different label proportions and categories on the three databases shown in Table~\ref{ablation_3datasets}. Our Resnet-50 method only uses the 2048D features representation extracted from the pre-trained ResNet-50 architecture without bilinear pooling and SAM operation. Our FBP is the method that uses bilinear pooling features representation. The method of SAM ResNet-50 uses the proposed self-boosting attention mechanism in ResNet-50, where the model does not use bilinear pooling features and only uses the last convolutional feature as the classifier's input. The method of SAM bilinear uses the proposed self-boosting attention mechanism in FBP.

From Table~\ref{ablation_3datasets}, we make the following observations: First, compared to the method of ResNet-50, SAM ResNet-50 increases the classification accuracy. For example, with the label proportion of $10\%$, $15\%$, $30\%$ and $50\%$ and the category of 200, the classification accuracy of SAM ResNet-50 are $40.24\%$, $52.05\%$, $64.07\%$ and $73.92\%$ respectively, which are $3.25\%$, $3.17\%$, $1.47\%$ and $0.69\%$ higher than ResNet-50 method on the \texttt{CUB200-2011} datasets. Similarly, significant improvement can also be found on the \texttt{Stanford Cars} and \texttt{FGVC Aircraft} datasets. This demonstrates the superiority of the proposed self-boosting attention mechanism. The model with a few label proportions is prone to overfit spurious correlated patterns. The proposed self-boosting attention mechanism regularizes the network and improves the classification performance in the testing set. A similar conclusion can also be found in comparing FBP and SAM bilinear. Second, compared to the method of ResNet-50, FBP has a better performance in most cases. This is because the method of FBP uses the bilinear pooling feature representation, which is more discriminative features for fine-grained visual recognition. Third, when the label proportion reduces from $50\%$ to $10\%$, the gap performances between SAM ResNet-50/SAM bilinear and ResNet-50/FBP become larger. Fourth, with the category of label reduced from 200 to 30 on the \texttt{CUB200-2011} datasets, 196 to 30 on the \texttt{Stanford Cars} dataset, and 100 to 30 on the \texttt{FGVC Aircraft} dataset, respectively, there is generally a better improvement in the performance of the proposed model. The proposed self-boosting attention mechanism effectively regularizes the network and reduces over-fitting under smaller label proportions or category labels, which is more beneficial for fine-grained visual recognition.
\subsection{Comparison with State-of-the-Art}
\label{CVS}
\begin{table}[!h]
\caption{Comparison with state-of-the-art FGVC methods with three label proportions on the three datasets. We also apply the proposed SAM to the state-of-the-art method DBTNet \cite{zheng2019learning}, creating a method dubbed SAM DBTNet-50, which shows compelling results with low feature dimension. }
\label{sota}
	\centering
	\resizebox{\textwidth}{45mm}{
\begin{threeparttable} 
\begin{tabular}{@{}c|l|c|ccccc@{}}
\toprule
\multirow{2}{*}{Dataset}       & \multirow{2}{*}{Method} & \multicolumn{1}{l|}{\multirow{2}{*}{Dimension $D$}} & \multicolumn{4}{c}{Label Proportion}                                                                               \\ \cmidrule(l){4-8} 
                               &                         & \multicolumn{1}{l|}{}                & 10\%               & 15\%                                 & 30\%                                 & 50\%            & 100\%                     \\ \midrule
\multirow{8}{*}{Bird}   & Fine-Tuning            & 2048                                         &36.99\%        & 48.88\%                              & 62.60\%                              & 73.23\%        &81.34\%                      \\
                               & FBP~\cite{lin2017bilinear}                      & 2048*16                                     & 37.88\%                                  & 49.12\%                                   & 63.27\%                                  & 73.70\%       &    82.52\%                     \\
                               & CBP-TS\cite{gao2016compact}                  & 500                                 &37.12\%              & 47.82\%                              & 62.24\%                              & 72.37\%      &    81.48\%                    \\
                               & HBP~\cite{yu2018hierarchical}                     & 8192*$\frac{n(n-1)}{2}$\dag                   &38.57\%                       & 50.12\%         & 63.86\%          & 74.18\%    &   86.12\%  \\
                               & DBTNet-50~\cite{zheng2019learning}                     & 2048                  &37.67\%                       & 49.52\%         & 63.16\%          &73.28 \%    &   86.04\%  \\ \cline{2-8}
                               & SAM ResNet-50       & 2048               & 40.24\%          & 52.05\%            & 64.07\%            & 73.92\% &  81.62\%        \\
                               & SAM DBTNet-50                    & 2048                                   &40.38\%        & 52.02\% &64.82\% & 74.12\% &\textbf{87.26\%}\\
                               & SAM bilinear                   & 2048*16                                   &\textbf{41.83\%}        & \textbf{52.35\%} &\textbf{65.19\%} & \textbf{74.54\%} &81.86\%\\ \midrule
\multirow{8}{*}{Car} & Fine-Tuning             & 2048                        &37.45\%                         & 53.01\%                              & 75.26\%                              & 83.56\%         &       91.02\%              \\
                               & FBP~\cite{lin2017bilinear}                      & 2048*16                        & 40.13\%                                  & 55.07\%                                  & 76.42\%                                  & 85.10\%       &   91.63\%                      \\
                               & CBP-TS\cite{gao2016compact}                  & 500                                   &37.77\%            & 54.87\%                              & 75.51\%                              & 84.80\%      &   89.52\%                     \\
                               & HBP~\cite{yu2018hierarchical}                     & 8192*$\frac{n(n-1)}{2}$\dag             &40.02\%                                 & 55.82\%         & 76.81\%          & 85.31\%     &  92.73\%   \\
                               & DBTNet-50~\cite{zheng2019learning}                     & 2048                  &39.48\%                       & 55.24\%         & 76.52\%          &  86.52\%   &   \textbf{94.32}\%  \\\cline{2-8}
                               & SAM ResNet-50       & 2048              & 39.96\%\            & 55.02\%            & 76.69\%            & 84.85\% &   91.06\%        \\
                               & SAM DBTNet-50                  & 2048                                   &42.47\%        & 56.06\% &\textbf{78.06\%} & \textbf{86.86\%}& 94.18\%\\
                               & SAM bilinear                     & 2048*16                          &\textbf{43.19\%}                 & \textbf{57.42\%} & 77.63\% & 85.71\% &91.48\%\\ \midrule
\multirow{8}{*}{Aircraft} & Fine-Tuning             & 2048                              &43.52\%                   & 53.17\%                              & 71.32\%                              & 78.61\%           &    87.13\%               \\
                               & FBP~\cite{lin2017bilinear}                      & 2048*16                            &45.16\%                                  & 55.06\%                                   & 72.12\%                                  & 79.93\%               & 87.32\%                \\
                               & CBP-TS\cite{gao2016compact}                  & 500                            &44.63\%                   & 54.79\%                              & 71.32\%                              & 79.60\%          &84.58\%                    \\
                               & HBP~\cite{yu2018hierarchical}                     & 8192*$\frac{n(n-1)}{2}$\dag                &45.28\%                             & 56.12\%          & 72.58\%          & 81.47\%     &89.74\%\\
                               & DBTNet-50~\cite{zheng2019learning}                     & 2048                  &45.35\%                       & 56.36\%         & 73.06\%          & 81.26\%    &   90.86\%  \\\cline{2-8}
                               & SAM ResNet-50           & 2048            & 46.73\%            & 56.02\%            & 72.59\%           & 79.21\%& 86.74\%       \\
                               & SAM DBTNet-50                    & 2048                                   &47.56\%        & \textbf{58.24\%} &73.36\% & \textbf{81.62\%} &\textbf{91.18\%}\\
                               & SAM bilinear                     & 2048*16                   &\textbf{47.97\%}                        & 57.47\% & \textbf{73.43\%} & 80.86\%        &87.46\%  \\ \bottomrule
\end{tabular}
\begin{tablenotes}  
\footnotesize              
\item [\dag] $n$ is the number of convolution layers features.          
\end{tablenotes}
\end{threeparttable} } 
\end{table}
We compare our method with state-of-the-art bilinear pooling methods on the \texttt{CUB200-2011}, \texttt{Stanford Cars} and \texttt{FGVC Aircraft} datasets shown in the Table~\ref{sota}. We can see that our SAM-based methods achieve state-of-the-art accuracy on the few label proportions on all these fine-grained datasets. Especially, we more significantly improve the classification accuracy on $10\%$ and $15\%$ label proportions compared to the improvement in $30\%$, $50\%$ and $100\%$ label proportions. We also incorporate the proposed SAM into the existing method of DBTNet-50~\cite{zheng2019learning}, which improves the performance when only a few annotations are available compared to the original DBTNet-50 method.

For the computational complexity, the bilinear feature dimensions in the method of FBP~\cite{lin2017bilinear} in our experiments is $2048*16$. The method of CBP-TS \cite{gao2016compact} is proposed to reduce the bilinear feature dimension. Setting the reduced dimension as 500 in our experiments can achieve the best performance. The dimension on the method of HBP~\cite{yu2018hierarchical} is 8192*$\frac{n(n-1)}{2}$ where 8192 is the embedding dimension obtained by the project layer in \cite{yu2018hierarchical}, and $n$ is the number of convolution layers features. We can find that with the increase of $n$, the dimension of the bilinear feature is higher. The dimension on the method of DBTNet-50~\cite{zheng2019learning} is 2048 to keep feature dimensions unchanged. In our method, the proposed SAM can be used in DBTNet-50, demonstrating that SAM DBTNet-50 has a better performance when only a few annotations are available than the original DBTNet-50. In the proposed SAM bilinear, the dimension of bilinear pooling features in our method is $2048*16$, ensuring an acceptable dimension and high classification performance. In the proposed SAM ResNet-50, the feature dimension is unchanged and equal to 2048, resulting in large computation cost savings.
\subsection{Analysis of the Number of Linear Projections in SAM-Bilinear}
\label{attentionnumber}
\begin{figure}[!htbp]
	\centering
	\includegraphics[width=3in]{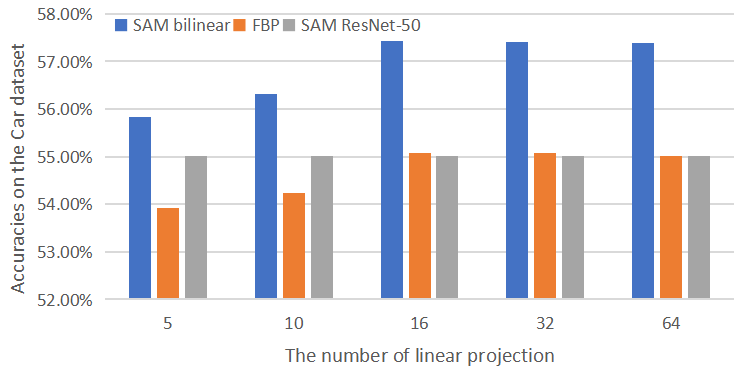}
	\caption{Comparison of SAM bilinear, FBP and SAM ResNet-50 with different number of linear projections on the Stanford Cars dataset.}
	\label{Numatt}
\end{figure}
\begin{figure}[tbp]
	\centering
	\includegraphics[width=3in]{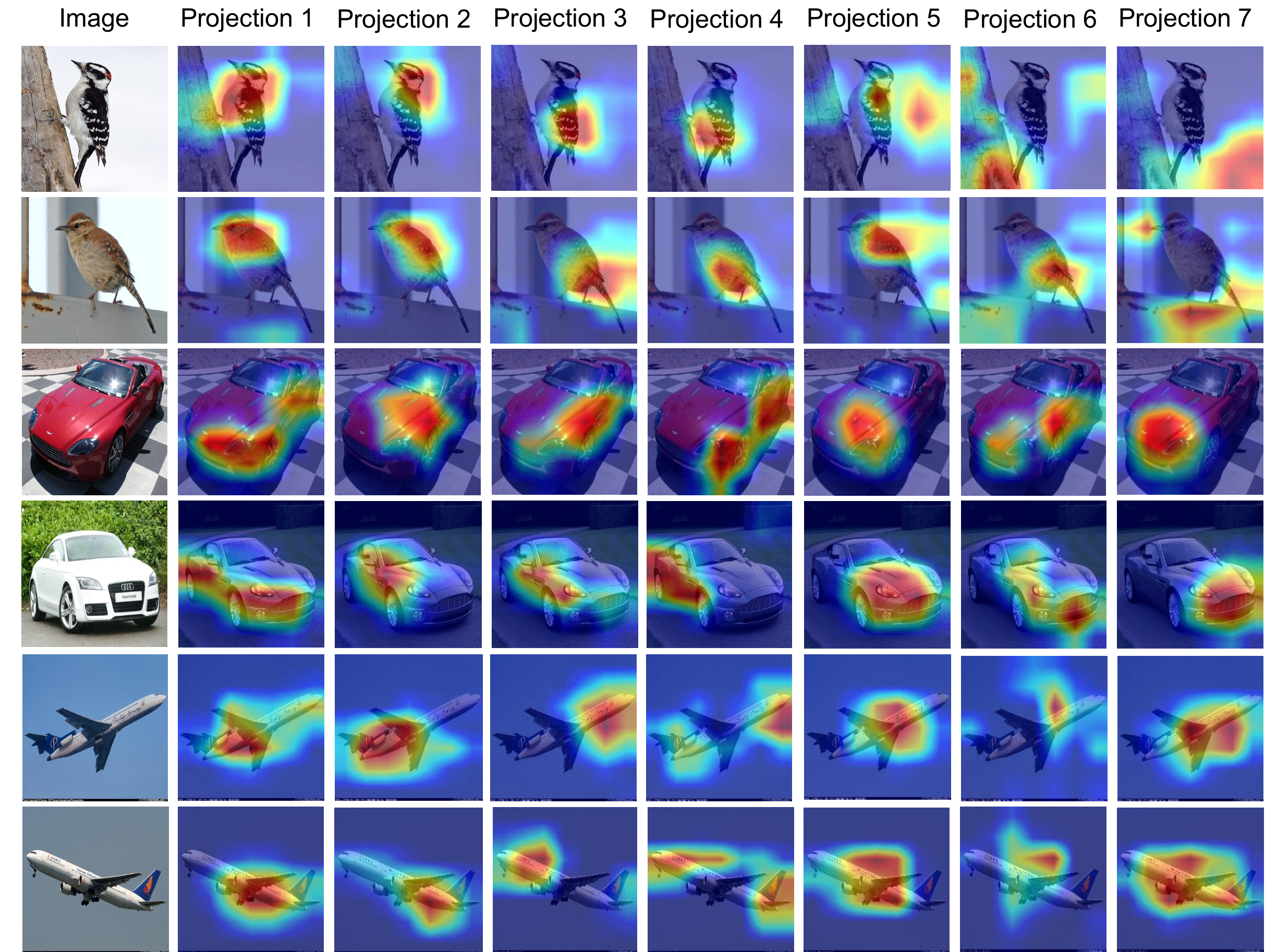}
	\caption{Visualization of each part detector in the multi-projection on the three datasets. The first column is the original input images. The 2-8 columns are the visualization of the seven detected attention regions in seven linear projections.}
	\label{visual1}
\end{figure}
\begin{figure}[htbp]
	\centering
	\includegraphics[width=3in]{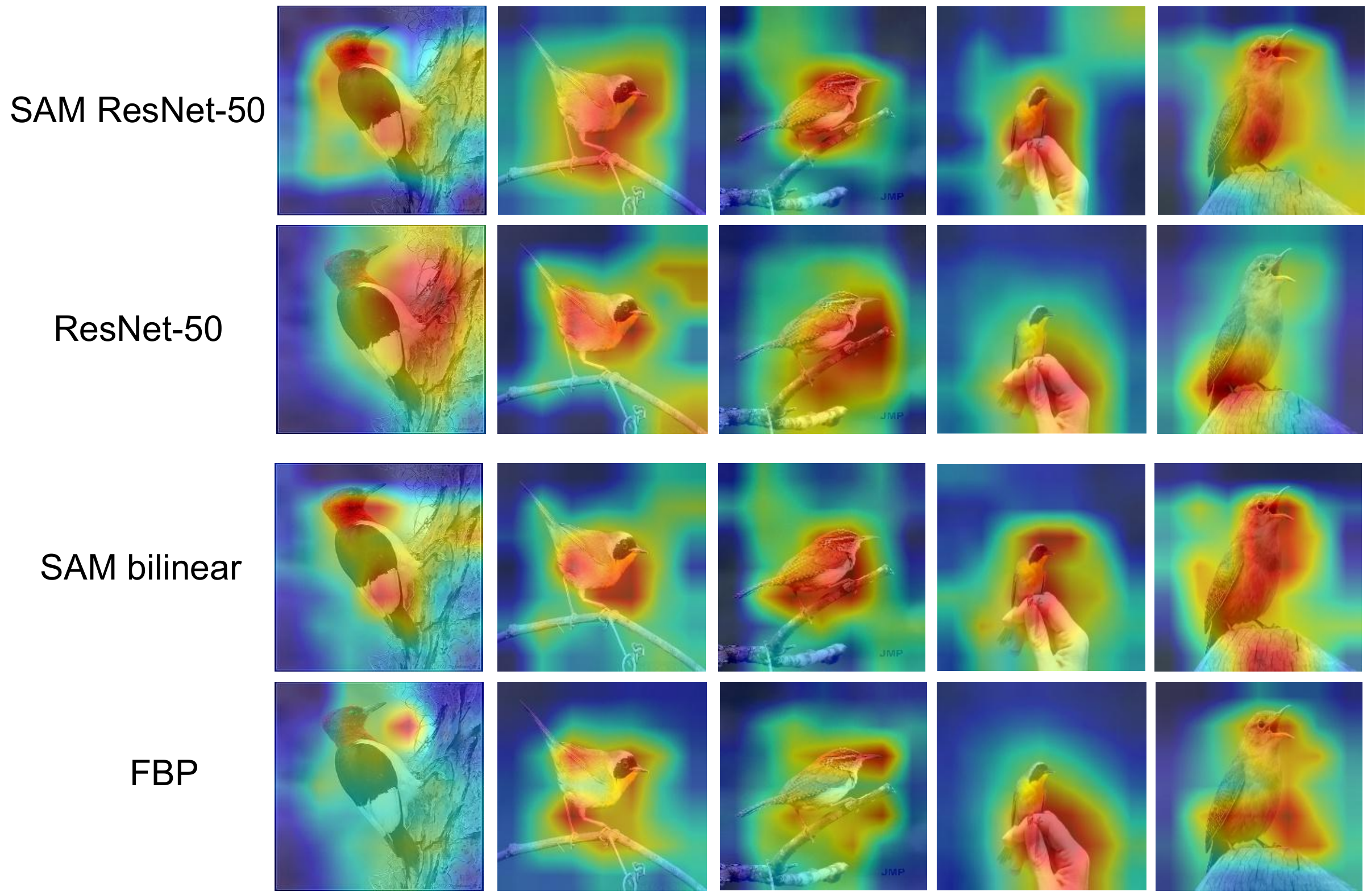}
	\caption{Visualization of the attention regions in the method of SAM ResNet-50 vs ResNet-50 and SAM bilinear vs FBP with $15\%$ label proportion.}
	\label{visual2}
\end{figure}
As elaborated in Section~\ref{extension_sam_bilinear}, we use multiple projections to leverage bilinear pooling operations to obtain a new representation of the image. It is also vital and can be used as part detectors to help the proposed network locate the object's discriminative part. To explore the impact of linear projections number $K$, we conduct experiments on the proposed SAM bilinear, FBP and SAM ResNet-50 by setting the different numbers of linear projections. Take the Stanford Cars datasets with $15\%$ label proportion, for example, our classification accuracy w.r.t. five different projections numbers are shown in Figure~\ref{Numatt}. From Figure~\ref{Numatt}, the accuracy significantly increases then gradually becomes stable in the method of SAM bilinear and FBP. The accuracy is peaked around 16, then slowly decrease with more heads (but only slightly). Please note that SAM ResNet-50 only needs one linear projection, and thus its accuracy is a constant in Figure~\ref{Numatt}.   
\subsection{Visualization}
\label{Vis}
\subsubsection{Visualization of Each Linear Projection}
The multi-projection has some practical implications. In our method, the number of linear projections is 16, and we visualize each result of linear projection under $15\%$ label proportion and show them partly in Figure~\ref{visual1}. As we can see, the highlighted regions of multi-projection reveal the significant parts that humans also rely on to improve the discriminative image representation, e.g., the head, body, and back for a bird, the head, tire and light for cars, and wings, head, and tail for aircraft.
\subsubsection{Visualization of Attention Regions for SAM and SAM-Bilinear}
This visualization aims to explain why the proposed method is effective when the number of training data becomes small. We compare the method of ResNet-50 with the proposed SAM ResNet-50, FBP with the proposed SAM bilinear, and visualize their attention regions on the \texttt{CUB200-2011} dataset with $15\%$ label proportions. From the visualization in Figure~\ref{visual2}, we can see that the existing method may not attend to the correct regions when the number of training samples becomes small. In contrast, the proposed methods, either SAM or SAM-Bilinear can produce a more reasonable attention map. 
This indicates that the self-boosting attention mechanism can be used to correct the predicted attention regions when the number of training data becomes smaller, thus improving the performance of the fine-grained visual recognition task.

\section{Conclusions}
In this paper, we propose a self-boosting attention mechanism (SAM) for fine-grained visual recognition to regularize the network with low data regimes. The proposed SAM enforces the network to focus on the key regions shared across samples and classes. These key regions are constrained to fit the attention maps generated from CAM/GradCAM. Unlike previous work identifying the key regions that rely on abundant training data, our self-boosting attention mechanism is still effective when the number of training samples becomes smaller. Furthermore, we extend the proposed SAM with the bilinear model to further strengthen the regularization. The proposed SAM effectively regularize the network when image-level annotations are quite a few, and outperforms existing state-of-the-art on the \texttt{CUB200-2011}, \texttt{Stanford Cars} and \texttt{FGVC Aircraft} datasets.
\clearpage
%
%
\bibliographystyle{splncs04}
\bibliography{egbib}
\end{document}